\documentclass[runningheads]{llncs}
\usepackage{makeidx}  
\usepackage{amsmath}
\usepackage{amssymb}
\usepackage{graphicx}
\usepackage{subfig}
\graphicspath{{figs/}}
\newcommand{\widthFour}{0.22\textwidth}
\usepackage{booktabs}
\makeatletter
\newcommand*\dashline{\rotatebox[origin=c]{90}{$\dabar@\dabar@\dabar@$}}
\makeatother
\begin{document}
\frontmatter          
\pagestyle{headings}  
\addtocmark{Sparse Projections of Medical Images onto Manifolds} 
\mainmatter              
\title{Sparse Projections of Medical Images\\ onto Manifolds}

\titlerunning{Sparse Projections of Medical Images onto Manifolds}  
%
\author{George H.~Chen \and
        Christian Wachinger \and
        Polina Golland}

\authorrunning{G.H.~Chen et al.} 

\institute{Massachusetts Institute of Technology,
           Cambridge MA 02139, USA \\
           \email{\{georgehc,wachinger,polina\}@csail.mit.edu}}

\maketitle              

\begin{abstract}
Manifold learning has been successfully applied to a variety of medical
imaging problems. Its use in real-time applications requires fast projection
onto the low-dimensional space. To this end, out-of-sample extensions are
applied by constructing an interpolation function that maps from the input
space to the low-dimensional manifold. Commonly used approaches such as the
Nystr\"{o}m extension and kernel ridge regression require using all training
points. We propose an interpolation function that only depends on a small
subset of the input training data. Consequently, in the testing phase each new
point only needs to be compared against a small number of input training data
in order to project the point onto the low-dimensional space. We interpret our
method as an out-of-sample extension that approximates kernel ridge
regression. Our method involves solving a simple convex optimization problem
and has the attractive property of guaranteeing an upper bound on the
approximation error, which is crucial for medical applications. Tuning this
error bound controls the sparsity of the resulting interpolation function. We
illustrate our method in two clinical applications that require fast mapping
of input images onto a low-dimensional space.
\end{abstract}

\section{Introduction}

Manifold learning maps high-dimensional data to a low-dimensional manifold
and has recently been successfully applied to a variety of applications.
Specifically in medical imaging, manifold learning has been used in
segmentation~\cite{zhang06}, registration~\cite{hamm09,rohdeWPM08},
computational anatomy~\cite{gerber09},
classification~\cite{BhatiaRPWHR12,wachinger2010mani},
detection \cite{suzuki2010massive}, and respiratory
gating~\cite{georg08manifold4D,wachinger2010miccai}. But to the best of our
knowledge, little work has been done using manifold learning for medical
imaging applications that require fast projections onto a low-dimensional
space.

In this paper, we demonstrate a method that achieves fast projection of input
data onto a low-dimensional manifold by constructing a projection function
that only depends on a small subset of the training data. Our method is a
sparse variant of kernel ridge regression~\cite{krr} and can be interpreted as
an interpolation function optimized to only use a few of the training data.
Furthermore, the construction of the interpolation function guarantees an
upper bound on an interpolation error for training data. The error is
measured in terms of the average squared Euclidean distance between the
predicted points of the interpolator versus those of kernel ridge regression
using all the points. As our interpolator has no parametric model for the data
points, its complexity is driven by the complexity of the training data and
the bound on the approximation error.

\textbf{Related work on out-of-sample extensions.}
Manifold learning is a specific case of nonlinear dimensionality reduction and
refers to a host of different algorithms
\cite{dimensionality_reduction_review}. In medical image analysis, manifold
learning is used to construct a low-dimensional space for images in which
subsequent statistical analysis (regression, classification, etc.) is
performed. Many manifold learning techniques do not construct a mapping of the
entire input space but only of the training points. For these methods,
estimating a new point's location in the low-dimensional space is performed
via an out-of-sample extension \cite{out_of_sample_ext}, with Nystr\"{o}m
extensions commonly used. For certain manifold learning methods, a Nystr\"{o}m
extension is a special case of kernel ridge
regression~\cite{learning_with_kernels}, and for both the Nystr\"{o}m
extension and kernel ridge regression, the resulting interpolation function
for mapping a new input point to the low-dimensional space depends on all
training data. Thus, we need to compare a new point to all training data
points, which is computationally expensive for volumetric images, especially
if the number of input data used to learn the manifold is large.

Our work is most similar to reduced rank kernel ridge regression~\cite{rrkrr},
which also approximates kernel ridge regression by only using a small number
of input training points. Reduced rank kernel ridge regression greedily
selects training points to minimize a particular cost function. Specifically,
the algorithm incrementally adds a training point that causes the largest
decrease in overall cost. Different criteria could be used for when the greedy
procedure is terminated such as if a pre-specified desired number of training
points to use is reached or if the overall cost drops below a pre-specified
desired error tolerance. Importantly, for medical applications, the latter
criterion is more directly connected to the error analysis of the whole
processing pipeline. Our approach also requires the user to specify a desired
error tolerance but uses a different cost function. Rather than using a greedy
approach to select which training points to add, we solve a convex
optimization problem implied by our cost function. We remark that the proposed
cost function also differs from that of support vector regression~\cite{svr},
which essentially achieves sparsity via excluding training points that map
sufficiently close to the estimated function. Our cost is more lenient, asking
that an average error be small rather than asking that an error be small for
each individual training point.

\textbf{Contributions.}
For high-dimensional input points $x_1,x_2,\dots,x_n\in\mathbb{R}^d$ and their
low-dimensional representations $y_1,y_2,\dots,y_n\in\mathbb{R}^p$ as computed
by any manifold learning algorithm, we propose a convex program for
constructing an out-of-sample extension that guarantees a bound on the
approximation error. Formally, if
$\widehat{f}:\mathbb{R}^d\rightarrow\mathbb{R}^p$ is the out-of-sample
extension function estimated via kernel ridge regression, then the sparse
projection function
$\widetilde{f}:\mathbb{R}^d\rightarrow\mathbb{R}^p$ constructed by our
algorithm satisfies
\begin{equation}
\frac{1}{n}
\sum_{i=1}^n
  \|\widehat{f}(x_i) - \widetilde{f}(x_i)\|_2^2
 \le \varepsilon^2,
\label{eq:main-guarantee}
\end{equation}
where $\|\cdot\|_2$ denotes the Euclidean norm,
$\varepsilon>0$ is a pre-specified error tolerance, and $\widetilde{f}$
depends only on a small subset of $x_1,\dots,x_n$. The size of the subset,
i.e., the sparsity of the resulting function~$\widetilde{f}$, depends on
tolerance $\varepsilon$ and training pairs $(x_1,y_1),\dots,(x_n,y_n)$.
Finding the smallest such subset is NP-hard. We instead consider a convex
relaxation with sparsity induced by a mixed $\ell_1/\ell_2$ norm. While the
proposed sparse approximation to kernel ridge regression can be used more
generally for other multivariate regression tasks, we restrict our focus in
this paper to out-of-sample extensions for manifold learning.

We apply our method to two medical imaging applications that require a fast
projection onto a low-dimensional space. The first application is respiratory
gating in ultrasound, where we assign the breathing state to each ultrasound
frame during the acquisition in real-time. The second application is the
estimation of a patient's position in a magnetic resonance imaging (MRI)
scanner while the patient is being moved to a target location.

\section{Background}
\label{sec:background}

Our method builds heavily on kernel ridge regression~\cite{krr}, reviewed
below. We also briefly discuss the result that a Nystr\"{o}m extension is a
special case of kernel ridge regression under certain
conditions~\cite{learning_with_kernels}. As a consequence, our sparse
approximation to kernel ridge regression also contains a sparse approximation
to the widely used Nystr\"{o}m extension.

\textbf{Kernel ridge regression.}
Let $\mathbb{H}$ be a family of functions mapping $\mathbb{R}^d$
to~$\mathbb{R}$ such that $\mathbb{H}$ is a reproducing kernel Hilbert space
(RKHS)~\cite{rk_theory} with kernel function
$\mathbb{K}:\mathbb{R}^d\times\mathbb{R}^d\rightarrow\mathbb{R}$.
Given points $x_1,\dots,x_n\in\mathbb{R}^d$ and
${y_1,\dots,y_n\in\mathbb{R}^p}$, 
we assume
that there exists a function $f^*=(f^*_1,\dots,f^*_p)\in\mathbb{H}^p$ such
that for each $i$, we have
$y_i=f^*(x_i) + w_i$ for some noise term $w_i\in\mathbb{R}^p$. 
Kernel ridge regression seeks an estimate $\widehat{f}$ of function $f^*$
by solving
\begin{equation}
\widehat{f}
=\underset{(f_{1},\dots,f_{p})\in\mathbb{H}^p}{\text{argmin}}
   \sum_{j=1}^p
     \left\{
       \sum_{i=1}^n
         (Y_{ij}-f_j(x_i))^2
          + \lambda\|f_j\|_{\mathbb{H}}^2
     \right\},
\label{eq:kernel-ridge-regression-multivariate}
\end{equation}
where
matrix $Y\in\mathbb{R}^{n\times p}$ contains data point $y_i$ as its
$i$-th row,
constant $\lambda>0$ controls the amount of regularization,
and
$\|\cdot\|_{\mathbb{H}}$ is the norm induced by the inner product of
$\mathbb{H}$.
The solution of optimization
problem~\eqref{eq:kernel-ridge-regression-multivariate} is 
\begin{equation}
\widehat{f}(\cdot)
=\sum_{i=1}^n
   \mathbb{K}(\cdot,x_i)
   \widehat{\alpha}_i,
\label{equ:krrEmb}
\end{equation}
where $\widehat{\alpha}_i$ refers to the $i$-th row of $n$-by-$p$ matrix
\begin{equation}
\widehat{\alpha}
=
\left( K+\lambda I_{n\times n} \right)^{-1} Y,
\label{eq:krr-multivariate-alpha-soln}
\end{equation}
matrix $K\in\mathbb{R}^{n\times n}$ is given by
$K_{ij}=\mathbb{K}(x_i,x_j)$, and $I_{n\times n}$ is the {$n\text{-by-}n$}
identity matrix~\cite{krr}.

\textbf{Nystr\"{o}m extension.}
The Nystr\"{o}m method approximates a certain type of eigenfunction problem
and is used for out-of-sample extensions in manifold
learning~\cite{out_of_sample_ext}. For manifold learning algorithms that
assign the low-dimensional coordinates directly from the eigenvectors of $K$,
e.g., Isomap~\cite{isomap}, locally linear embeddings~\cite{lle}, and
Laplacian eigenmaps~\cite{laplacian_lle}, we can derive the Nystr\"{o}m
extension as a special case of kernel ridge regression with $\lambda=0$.
Specifically, with eigendecomposition $K=\Phi\Lambda\Phi^{-1}$, where
$\Lambda=\text{diag}(\lambda_1,\lambda_2,\dots,\lambda_n)$ and
$\lambda_1\ge\lambda_2\ge\cdots\ge\lambda_n$, we consider when the
low-dimensional embedding is given by $Y=\Phi_\ell$, the matrix consisting of
the first $\ell$ columns of $\Phi$. If we use~$\phi^{(j)}$ to denote the
$j$-th column of $\Phi$, then with $\lambda=0$ and $Y=\Phi_\ell$,
eq.~\eqref{eq:krr-multivariate-alpha-soln} reduces to
\begin{equation}
\widehat{\alpha}
= K^{-1} Y
= \Phi \Lambda^{-1} \Phi^{-1} \Phi_\ell
= \Phi \Lambda^{-1}
  \begin{bmatrix}
    I_{\ell\times\ell} \\
    \mathbf{0}
  \end{bmatrix}
= \begin{bmatrix}
    \frac{1}{\lambda_1}\phi^{(1)}
    \dashline
    &
    \frac{1}{\lambda_2}\phi^{(2)}
    \dashline
    &
    \cdots
    \dashline
    &
    \frac{1}{\lambda_\ell}\phi^{(\ell)}
  \end{bmatrix}.
\label{eq:nystrom-alpha}
\end{equation}
Letting $\phi_i^{(j)}$ refer to the $i$-th element of $\phi^{(j)}$, and
substituting eq.~\eqref{eq:nystrom-alpha} into eq.~\eqref{equ:krrEmb}, we see
that, for a new point $x\in\mathbb{R}^d$, the $j$-th element of
$\widehat{f}(x)$ is given by
\begin{equation}
\widehat{f}_j( x )
=\sum_{i=1}^n
   \mathbb{K}( x ,x_{i}) \widehat{\alpha}_{ij}
=\sum_{i=1}^n
   \mathbb{K}( x ,x_{i})
   \left(
     \frac{1}{\lambda_j}\phi_i^{(j)}
   \right)
=\frac{1}{\lambda_j}
 \sum_{i=1}^n
   \phi_i^{(j)}
   \mathbb{K}( x ,x_{i}),
\end{equation}
which is the formula for the low-dimensional embedding of $x$ using the
Nystr\"{o}m extension \cite{out_of_sample_ext}. Importantly, kernel
function~$\mathbb{K}$ depends on the choice of a manifold learning algorithm
\cite{out_of_sample_ext}. The above relationship shows that for certain
manifold learning algorithms, kernel ridge regression is a richer model for
out-of-sample extensions than the Nystr\"{o}m extension.

\section{Sparse Approximation to Kernel Ridge Regression}
\label{sec:svkrr}

We now present our method. We seek an interpolation function
$\widetilde{f}:\mathbb{R}^d\rightarrow\mathbb{R}^p$ within a family of
functions
$\mathbb{G}=\{f(\cdot)=\sum_{i=1}^n \mathbb{K}(\cdot,x_i)\alpha_i
              : \alpha\in\mathbb{R}^{n\times p}\}$,
with many vectors $\alpha_i\in\mathbb{R}^p$ equal to zero while ensuring that
upper bound~\eqref{eq:main-guarantee} holds. In particular, we formulate a
convex optimization problem where $\alpha\in\mathbb{R}^{n\times p}$ is the
only decision variable; solving this problem yields $\widetilde{\alpha}$ that
implies a sparse approximation $\widetilde{f}$ to the kernel ridge regression
solution~$\widehat{f}$.

Because we optimize over functions in~$\mathbb{G}$, upper bound
\eqref{eq:main-guarantee} can be simplified by noting that
${\sum_{i=1}^n
    \|\widehat{f}(x_{i})-f(x_{i})\|_2^2}
 ={\|K\widehat{\alpha}-K\alpha\|_F^2}$,
where $\|\cdot\|_F$ denotes the Frobenius norm, and $\widehat{\alpha}$ is
given by eq.~\eqref{eq:krr-multivariate-alpha-soln}. In fact,
$\widehat{f}(x_i)$ and $f(x_i)$ are given by the $i$-th rows of
$K\widehat{\alpha}$ and $K\alpha$, respectively. Thus,
bound~\eqref{eq:main-guarantee} can be rewritten as
$\|K\widehat{\alpha}-K\alpha\|_F^2\le n\varepsilon^2$. Satisfying this
constraint while encouraging the number of nonzero vectors $\alpha_i$ to be
small can be achieved by solving the following convex optimization problem:
\begin{equation}
\widetilde{\alpha}
=\underset{\alpha\in\mathbb{R}^{n\times p}}
          {\text{argmin}}
    \sum_{i=1}^n \|\alpha_i\|_2
 \qquad
 \text{s.t.}
 \qquad
 \|K\widehat{\alpha}-K\alpha\|_F^2\le n\varepsilon^2.
\label{eq:eps-discrep-opt}
\end{equation}
By minimizing the mixed $\ell_1/\ell_2$ norm of $\alpha$, we encourage each
vector $\alpha_i$ to either consist of all zeros or all non-zero entries
\cite{sparse_optimization}. Note that if $p=1$ and we instead ask for the
sparsest solution possible, then the objective function becomes the $\ell_0$
norm (i.e., the number of nonzero elements) of $\alpha$, and the optimization
problem itself becomes NP-hard~\cite{l0_nphard}.

To solve optimization problem~\eqref{eq:eps-discrep-opt}, we reduce it to
solving many instances of its unconstrained Lagrangian form for which there is
already a fast solver. Specifically, by Lagrangian duality and convexity,
solving optimization problem~\eqref{eq:eps-discrep-opt} is equivalent to
solving the dual problem
\begin{equation}
\max_{\xi \ge 0}
    \min_{\alpha\in\mathbb{R}^{n\times p}}
      \left\{
        \sum_{i=1}^n
          \|\alpha_i\|_2
        + \xi
          \big(
            \|K\widehat{\alpha}-K\alpha\|_F^2 - n\varepsilon^2
          \big)
      \right\} \\
=\sup_{\xi > 0}
   \xi
   \big[
     g(1/\xi) - n\varepsilon^2
   \big], \label{eq:concave-maximization}
\end{equation}
where $\xi$ is a Lagrange multiplier, and
\begin{equation}
g(\gamma)
=\min_{\alpha\in\mathbb{R}^{n\times p}}
   \left\{
     \|K\widehat{\alpha}-K\alpha\|_F^2
     +
     \gamma
     \sum_{i=1}^n
       \|\alpha_i\|_2
   \right\}.\label{eq:mixed-l1-l2-opt}
\end{equation}
For a fixed $\xi$, we can compute $g(1/\xi)$ efficiently using the fast
iterative shrinkage-thresholding algorithm (FISTA) \cite{fista}. Moreover,
from a standard result of Lagrangian duality, dual
problem~\eqref{eq:concave-maximization} maximizes a concave function, which in
this case is only over scalar variable $\xi$. Thus, we can efficiently solve
the right hand side of~\eqref{eq:concave-maximization} by making as many calls
to FISTA as needed to achieve the desired accuracy in estimating $\xi$. Given
the final estimated value $\widetilde{\xi}$ of $\xi$, we recover
solution~$\widetilde{\alpha}$ by seeking $\alpha\in\mathbb{R}^{n\times p}$
that yields
$g(1/\widetilde{\xi})$ in eq.~\eqref{eq:mixed-l1-l2-opt}.

Once the coefficient matrix $\widetilde{\alpha}$ is obtained, the
interpolation function $\widetilde{f}$ is uniquely defined:
\begin{equation}
\widetilde{f}(\cdot)
=\sum_{i=1}^n
   \mathbb{K}(\cdot,x_i)
   \widetilde{\alpha}_i.
\label{eq:sparse-interpolator}
\end{equation}
The number of nonzero $\widetilde{\alpha}_i\in\mathbb{R}^p$ vectors depends on
error tolerance $\varepsilon$, regularization parameter $\lambda$, the kernel
function $\mathbb{K}$, and the data itself. We refer to the data points $x_i$
corresponding to nonzero $\widetilde{\alpha}_i$ as \textit{support vectors}.
As we observe empirically in the next section, decreasing parameters
$\varepsilon$ and $\lambda$ each generally produce more support vectors used
in projection. This is not surprising: increasing $\varepsilon$ increases the
size of the feasible set in optimization problem \eqref{eq:eps-discrep-opt},
allowing for potentially more candidate solutions $\widetilde{\alpha}$.
Meanwhile, as $\lambda\rightarrow\infty$, the coefficient matrix
$\widehat{\alpha}$ for kernel ridge regression, defined in
eq.~\eqref{eq:krr-multivariate-alpha-soln}, approaches
$\widehat{\alpha}=\frac{1}{\lambda}Y$, which goes to 0 for large $\lambda$.
As a result, $\widetilde{\alpha}$ also gets pushed to 0.

We can choose the similarity kernel $\mathbb{K}$ to match the specific choice
of manifold learning algorithm used to embed the training data. This allows us
to provide a sparse approximation to the Nystr\"{o}m extension as discussed in
Section~\ref{sec:background}. Alternatively, our method is applicable to any
kernel $\mathbb{K}$, regardless of the manifold learning algorithm used for
training.

Lastly, we note that solving the convex program
\eqref{eq:concave-maximization} to obtain $\widetilde{\alpha}$ incurs an
offline, one-time cost. During testing, we use the resulting sparse
interpolator~\eqref{eq:sparse-interpolator} whose computational cost is
directly proportional to the number of support vectors. Our interpolator will
always be at least as fast to compute as that of kernel ridge regression that
uses all the training points as support vectors and corresponds to the
special case of our interpolator where $\varepsilon=0$.

\section{Results}
\label{sec:results}

We apply our sparse interpolator to synthetic data (a Swiss roll), respiratory
gating in ultrasound, and MRI classification. We report the number of support
vectors as a proxy for computational speed since wall-clock time is directly
proportional to the number of support vectors. Furthermore, the datasets we
use are still relatively small for the scenarios our method intends to
address, making wall-clock time for the experiments we run not reflective of
real use.
However, our empirical results suggest that our
method can work with larger datasets since the number of support vectors
scales not with the size of the training dataset but instead with the
complexity of the training data's low-dimensional embedding.

For synthetic data, we use
Hessian eigenmaps~\cite{hessian_lle} for manifold learning, which, to the best
of our knowledge, does not have a known Nystr\"{o}m extension. For the
two experiments on real data, we use Laplacian eigenmaps~\cite{laplacian_lle}
for manifold learning and construct our sparse interpolator using the same
kernel function as the one used for Laplacian eigenmap's Nystr\"{o}m
extension~\cite{out_of_sample_ext}:
\begin{equation}
\mathbb{K}(x, x')
= \frac{W(x, x')}
       {\sqrt{\sum_{i=1}^n W(x, x_i) \sum_{j=1}^n W(x', x_j)}},
\label{eq:laplacian-ose-kernel}
\end{equation}
where $W:\mathbb{R}^d\times\mathbb{R}^d\rightarrow\mathbb{R}_+$ is a heat
kernel given by $W(x,x')=e^{-\|x-x'\|_2^2 / t}$ if $\|x - x'\|_2\le\tau$ and
0 otherwise, for some pre-specified temperature $t$ and nearest-neighbor
threshold $\tau$ --- both parameters chosen based on the application of
interest. We can also find the $k$ nearest neighbors rather than defining
nearest neighbors to be within a ball of radius $\tau$. With kernel
function~\eqref{eq:laplacian-ose-kernel}, constructing our sparse interpolator
with $\lambda=0$ and $\varepsilon=0$ yields Laplacian eigenmap's Nystr\"{o}m
extension that uses all the training points. We do not use the same manifold
learning algorithm for all datasets; the choice of manifold learning algorithm
depends on the dataset and the application of interest.

\subsection{Synthetic Data}
\label{sec:synth}

\begin{figure}[t]
\vspace{-1em}
\centering
\subfloat[][\label{fig:swiss-roll-data}]{
\includegraphics[width=3.74cm, clip=true, trim=25mm 25mm 20mm 30mm]{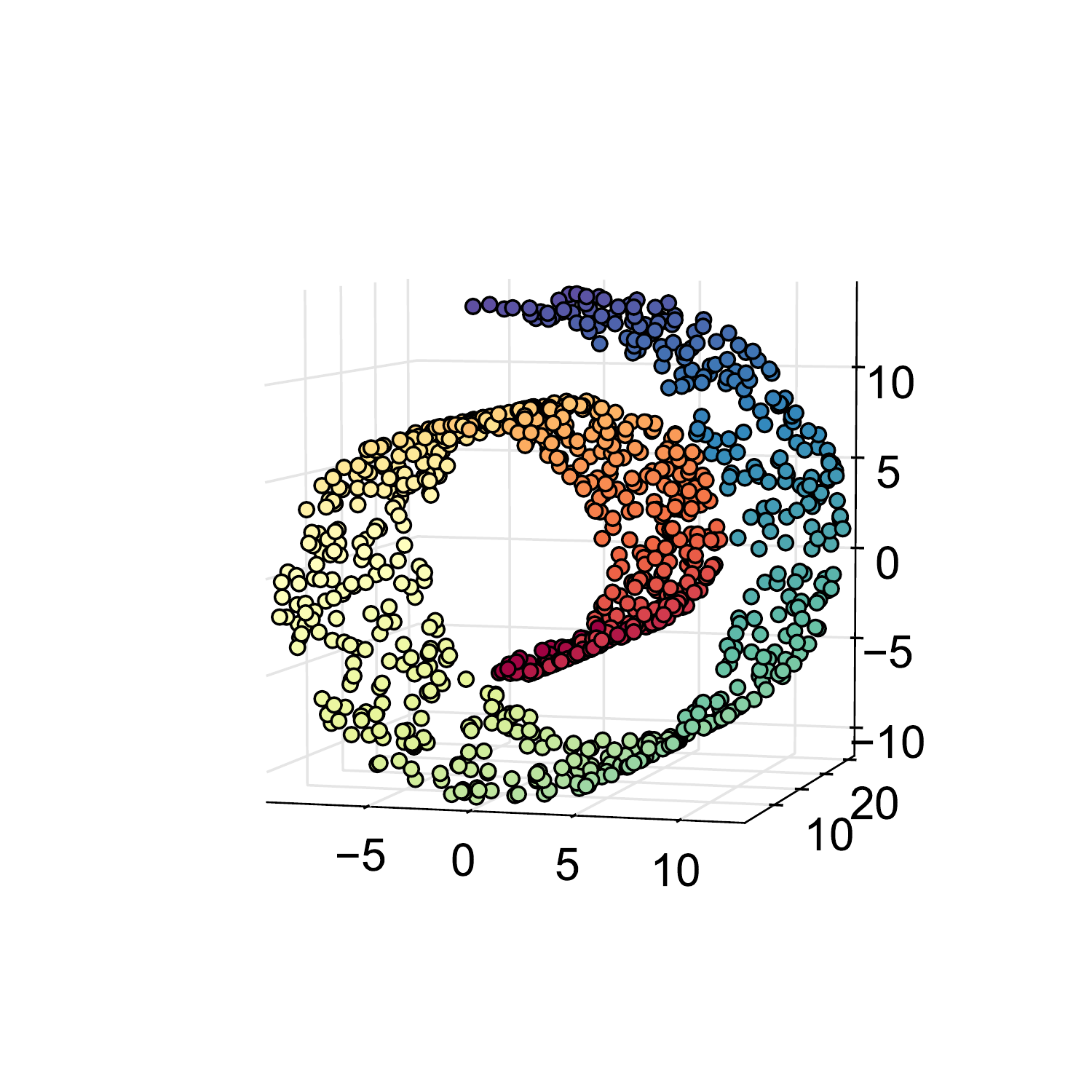}
}
~~
\subfloat[][\label{fig:swiss-roll-embedding}]{
\includegraphics[width=3.74cm, clip=true, trim=0mm 8mm 10mm 10mm]{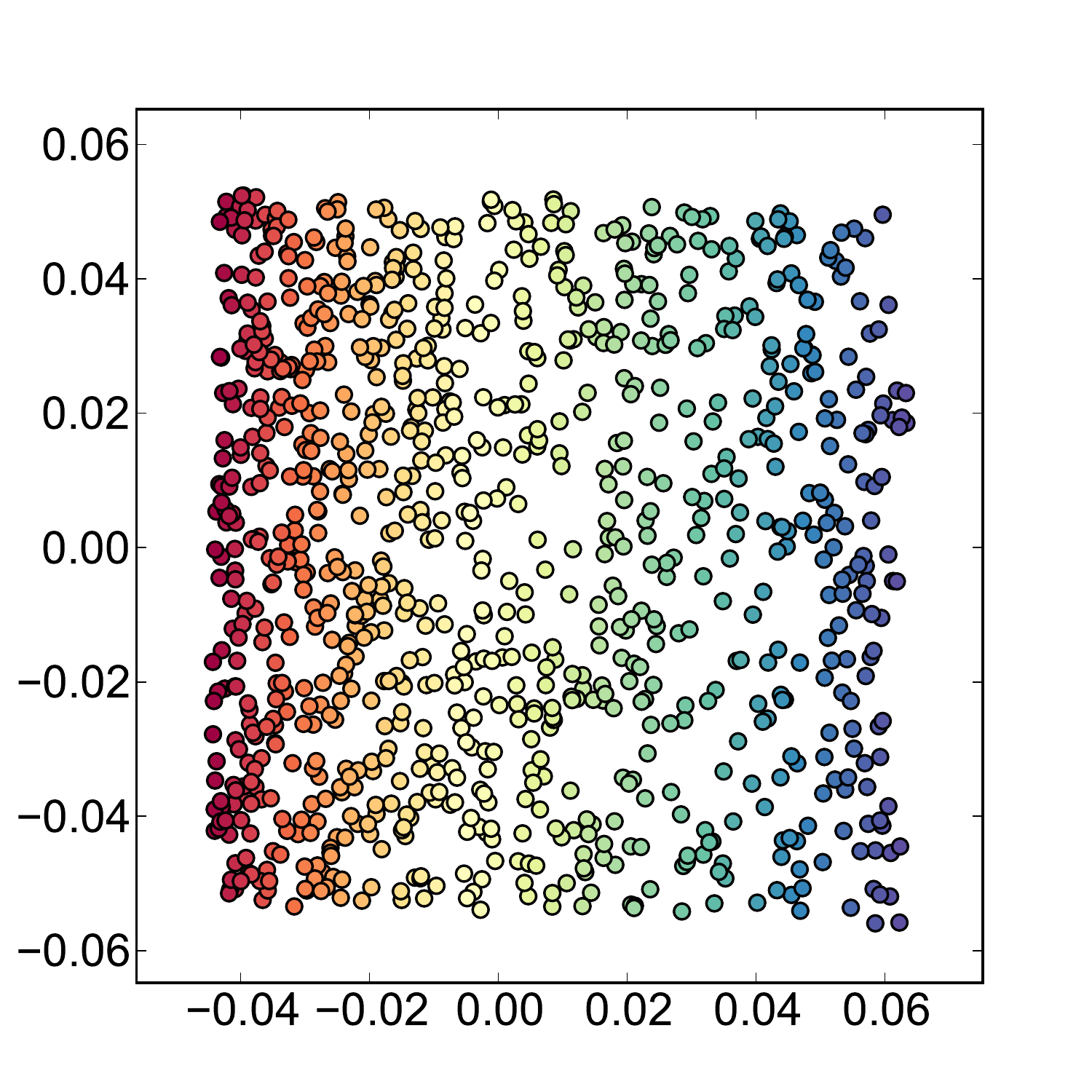}
}
~~
\subfloat[][\label{fig:swiss-roll-sparsity-vs-eps}]{
\includegraphics[width=3.74cm, clip=true, trim=3mm 4mm 1mm 3mm]{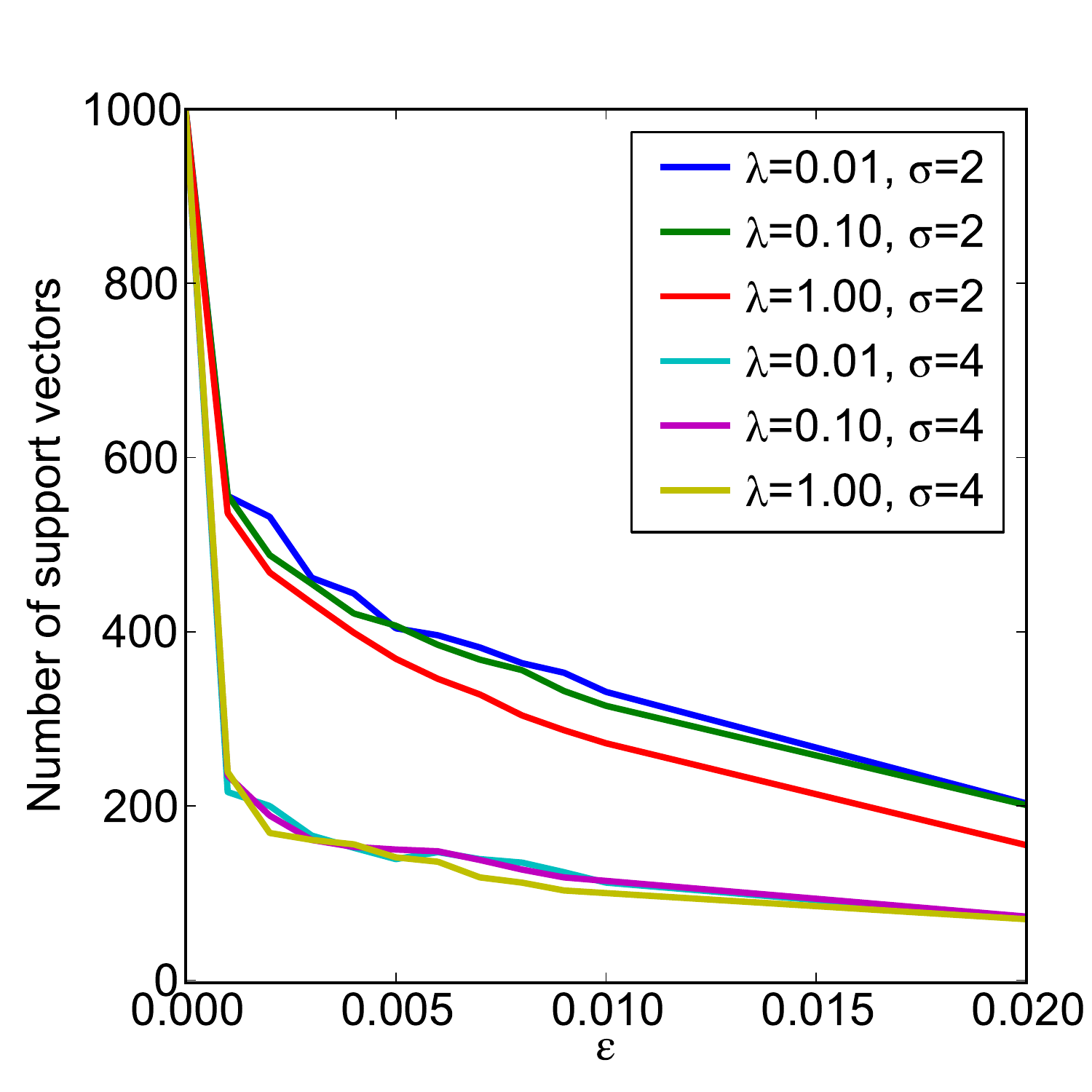}
}
\vspace{-1em}\\
\subfloat[][\label{fig:swiss-roll-data-sparse}]{
\includegraphics[width=3.74cm, clip=true, trim=25mm 25mm 20mm 30mm]{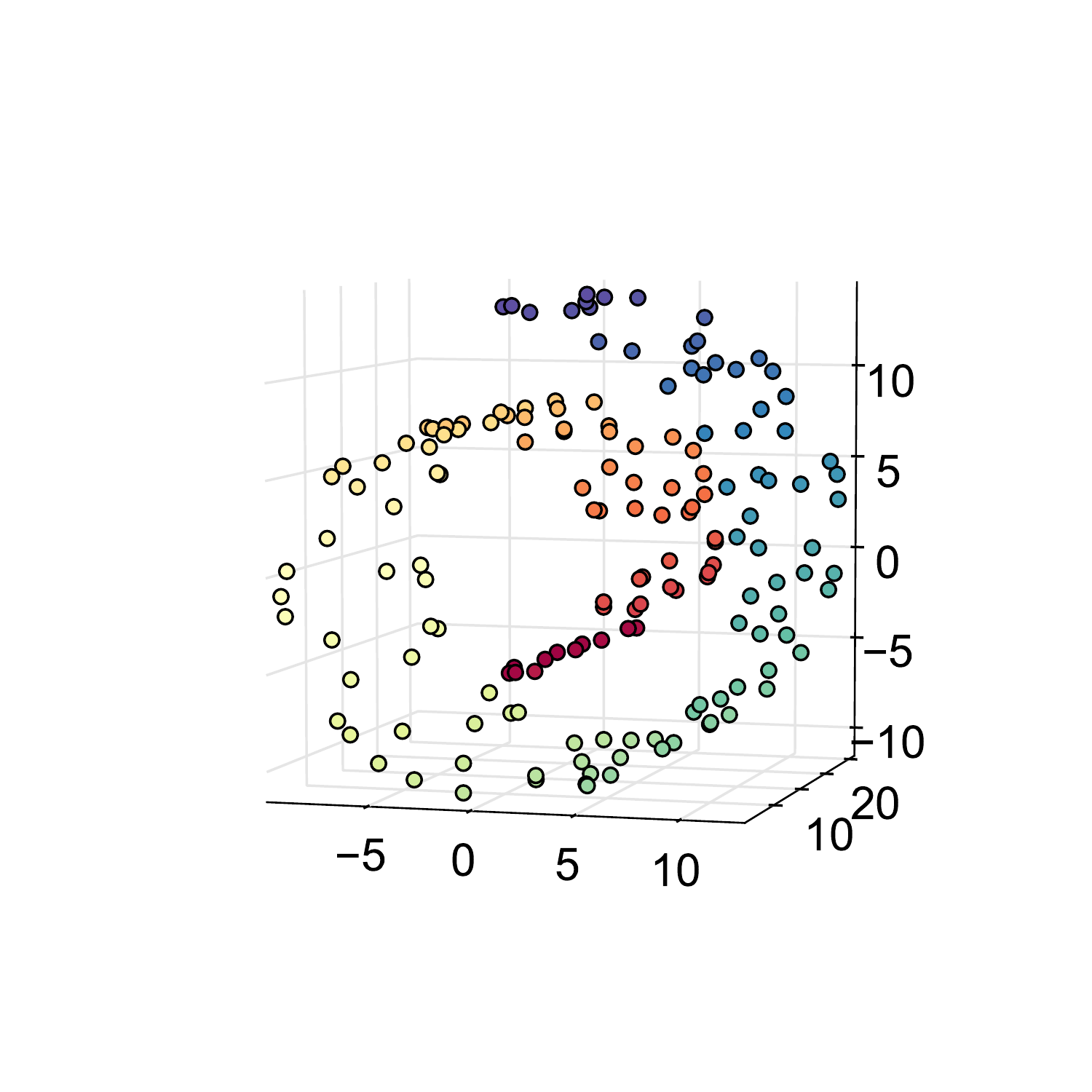}
}
~~
\subfloat[][\label{fig:swiss-roll-embedding-sparse-interp}]{
\includegraphics[width=3.74cm, clip=true, trim=0mm 8mm 10mm 10mm]{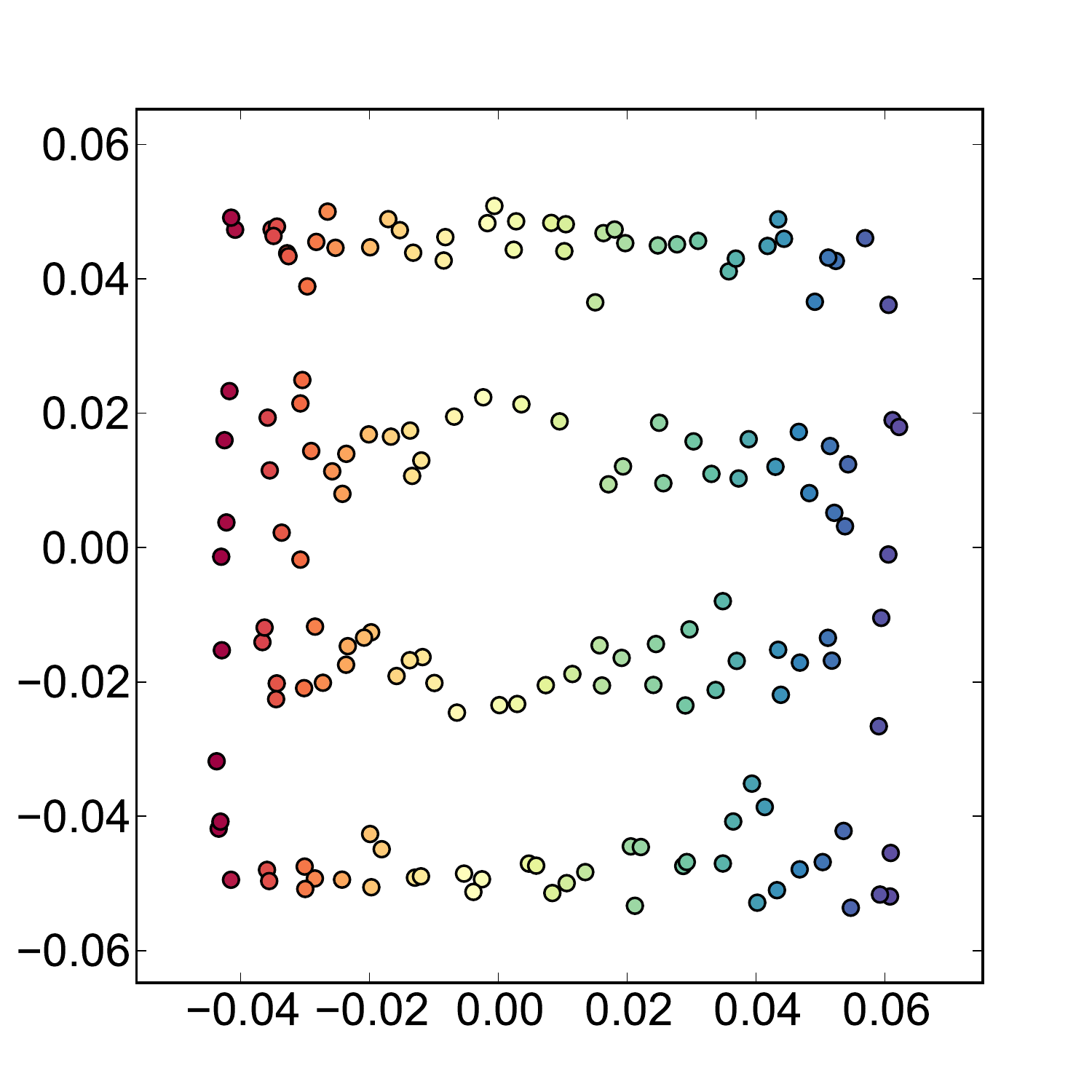}
}
~~
\subfloat[][\label{fig:swiss-roll-comparison}]{
\includegraphics[width=3.74cm, clip=true, trim=0mm 8mm 10mm 10mm]{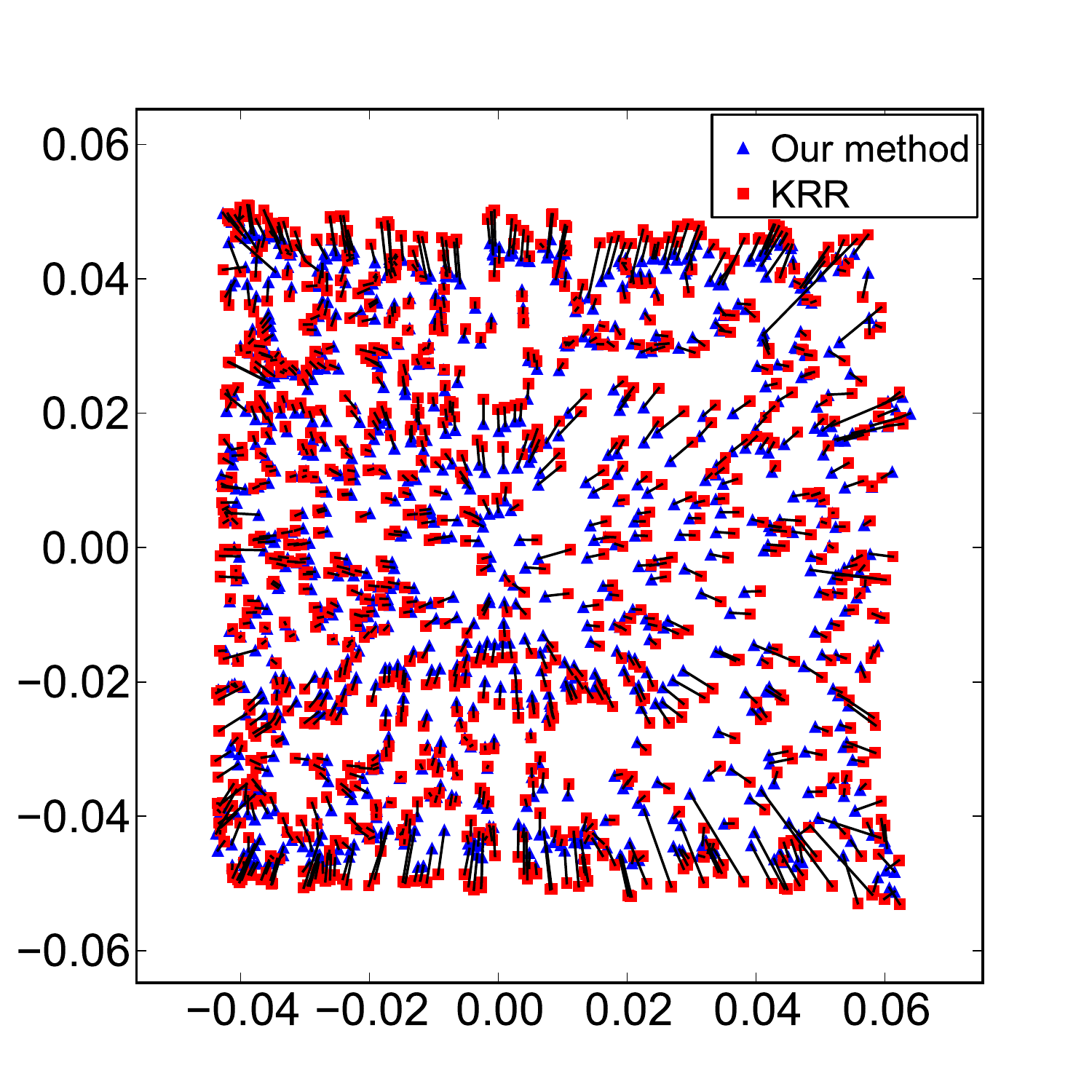}
}
\caption{Results for a Swiss roll with $n=1000$ points:
         (a) the original 3D data points;
         (b) their 2D embedding;
         (c) the number of support vectors as a function of error tolerance
             $\varepsilon$ for various $\lambda$ and $\sigma$.
         For the remaining panels (d)-(f), we fix
         $\lambda=0.1$, $\sigma=4$, and $\varepsilon=0.003$:
         (d) the 161 support vectors found;
         (e) our approximated 2D embedding of support vectors;
         (f) a comparison of 2D embeddings from our method and
             kernel ridge regression (lines show correspondences).
        }
\label{fig:swiss-roll}
\end{figure}

We apply our method to a Swiss roll with $n=1000$ points, shown in
Fig.~\ref{fig:swiss-roll}\subref{fig:swiss-roll-data}. First, we compute
low-dimensional representations $y_1,\dots,y_n\in\mathbb{R}^2$ using Hessian
eigenmaps~\cite{hessian_lle} with a 7-nearest-neighbor graph. We construct our
sparse interpolator using kernel function
$\mathbb{K}(x,x')=\exp(-\|x-x'\|_2^2/\sigma^2)$. To probe the behavior of our
interpolator, we vary kernel ridge regression parameter $\lambda$, kernel width
$\sigma$, and error tolerance $\varepsilon$. Fig.~\ref{fig:swiss-roll} reports
the resulting number of support vectors and illustrates results for one
setting of the parameters.

We observe that the support vectors are not uniformly sampled in the input
space nor on the learned 2D manifold. Instead, they appear along the
boundaries or form a skeleton within the learned manifold. We also observe in
Fig.~\ref{fig:swiss-roll}\subref{fig:swiss-roll-comparison} that the largest
discrepancies in the predicted point locations between our sparse interpolator
and kernel ridge regression occur along the boundaries. Unsurprisingly,
increasing kernel width~$\sigma$ reduces the number of support vectors needed
to achieve the same error tolerance~$\varepsilon$ as each support
vector has broader spatial influence in the input space. Furthermore,
increasing kernel ridge regression regularization parameter~$\lambda$ also
reduces the number of support vectors, as discussed in
Section~\ref{sec:svkrr}.

By repeating this experiment using a Swiss roll with $n=2000$, $n=3000$, and
$n=4000$ points, we empirically find that for a variety of parameter
settings $\lambda$, $\sigma$, and $\varepsilon$, the number of support vectors
remains roughly constant as~$n$ grows large. For example, with $\lambda=0.1$,
$\sigma=4$, and $\varepsilon=0.003$, we obtain 161, 174, 163, and 170 support
vectors for $n=1000,2000,3000,4000$ points respectively. This suggests that
the number of support vectors to depend on the low-dimensional embedding's
complexity and not on the dataset size $n$.

\subsection{Respiratory Gating of Ultrasound Images}
\label{sec:us}

\begin{figure}[t]
\begin{center}
\includegraphics[width=\widthFour]{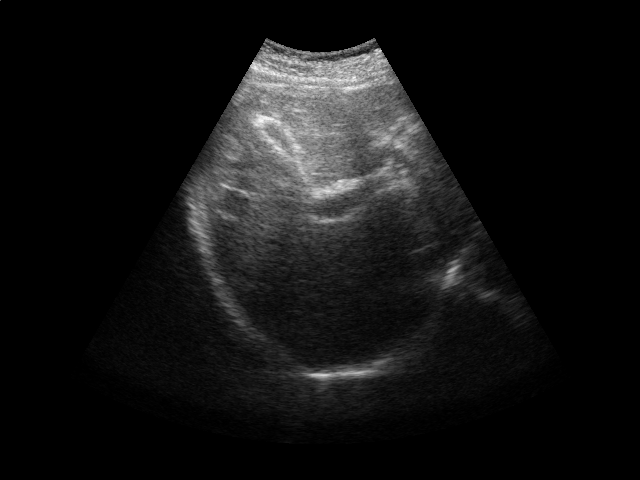}
\includegraphics[width=\widthFour]{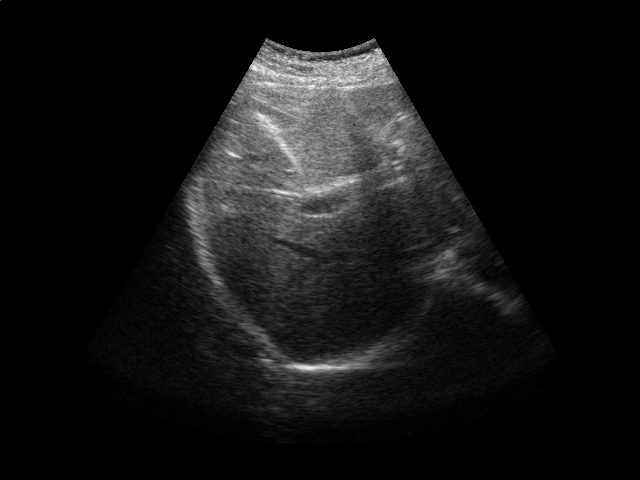}
\includegraphics[width=\widthFour]{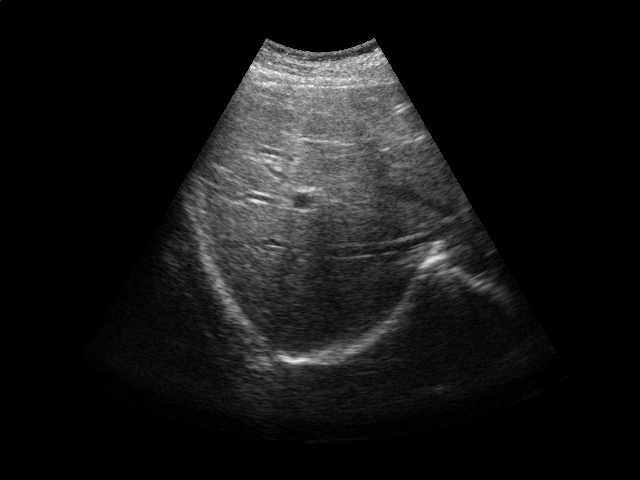}
\includegraphics[width=\widthFour]{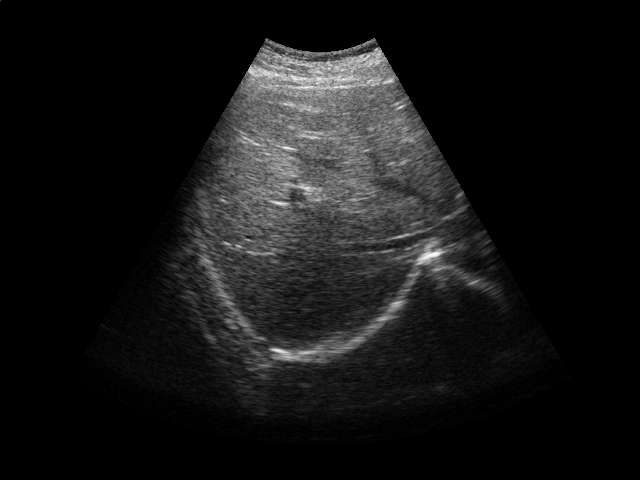}

~\\

\includegraphics[width=0.45\linewidth, clip=true, trim=.4in 2.56in .4in 2.5in]{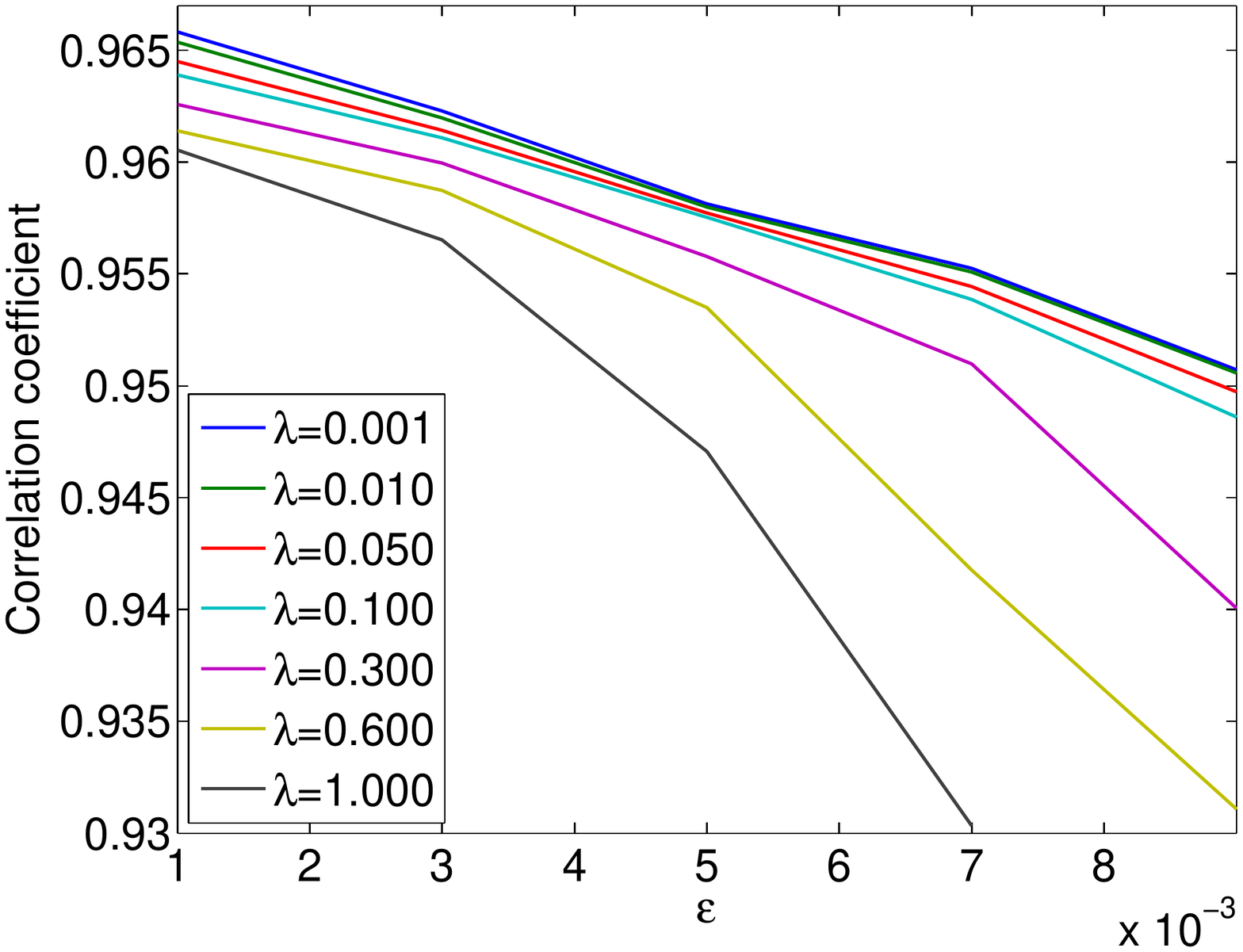}~~~~
\includegraphics[width=0.45\linewidth, clip=true, trim=.4in 2.56in .4in 2.5in]{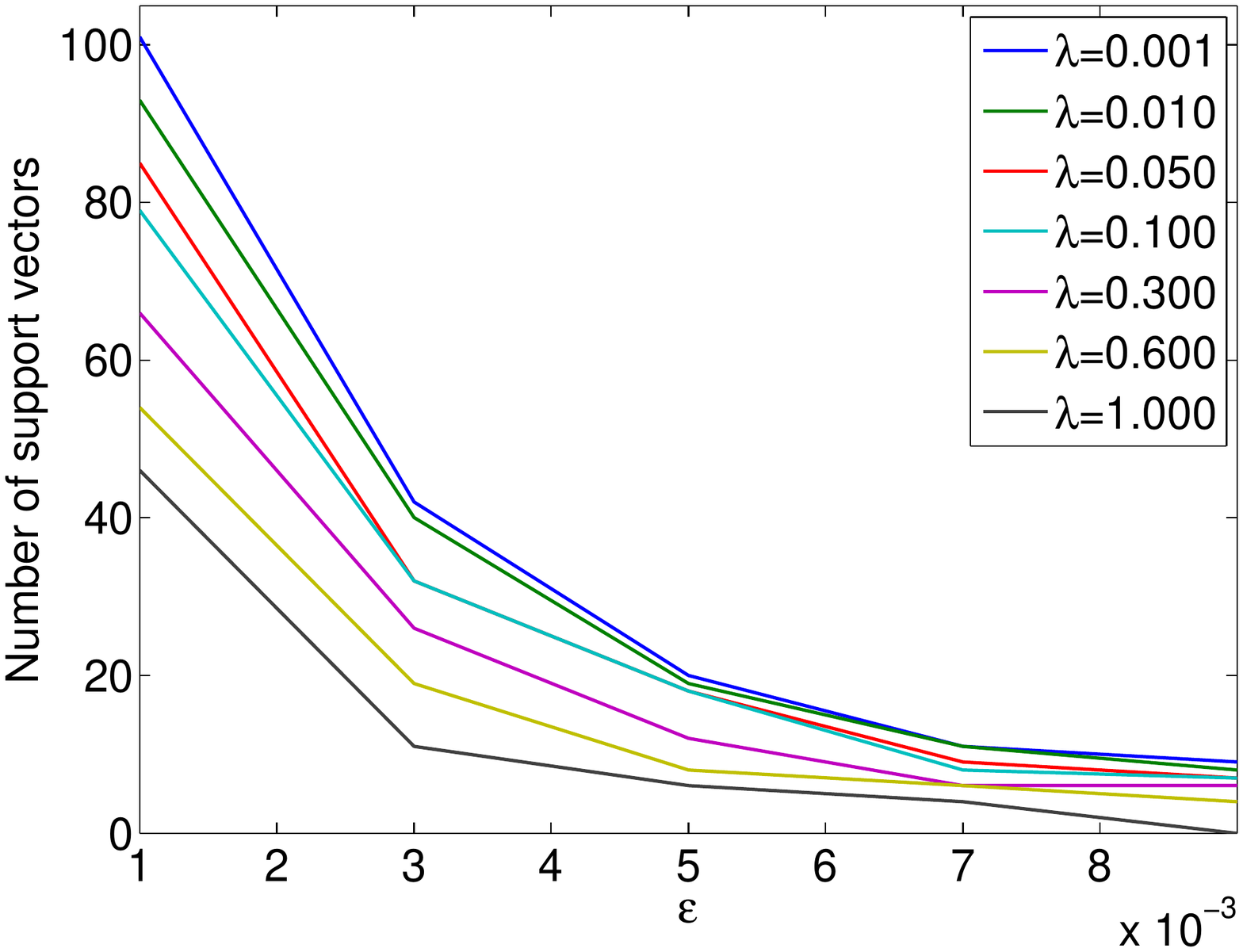}
\caption{Ultrasound gating.
         Top: Ultrasound images of the liver over time
         (abdomen, right upper quadrant).
         Bottom left: Correlation coefficient
         vs.~error tolerance~$\varepsilon$.
         Bottom right: The number of support vectors vs.~error
         tolerance~$\varepsilon$.
         Both figures in the bottom report results for different values of
         kernel ridge regression regularization parameter $\lambda$.
\label{fig:USimages}}
\end{center}
\end{figure}

\begin{table}[t]
\begin{center}
\footnotesize
\begin{tabular}{|c|c|ccc|ccc|}
\hline
& & \multicolumn{3}{c}{Learning on first 200 frames} & \multicolumn{3}{|c|}{Learning on entire data} \\
Data         & \hspace{1bp}\# Frames\hspace{1bp} &
   \hspace{1bp}CC (KRR)\hspace{1bp}   &  CC (sparse)\hspace{1bp} & \# SV's\hspace{1bp}  &
   \hspace{1bp}CC (KRR)\hspace{1bp}   &  CC (sparse)\hspace{1bp} & \# SV's\hspace{1bp}   \\
\hline
\hspace{1bp}Seq.~1\hspace{1bp}  &  354  &  96.5\%  &  96.4\%  &  79  &  99.9\%  &  96.9\%  &  73 \\
\hspace{1bp}Seq.~2\hspace{1bp}  &  335  &  97.7\%  &  97.5\%  &  99  &  99.9\%  &  98.6\%  & 100 \\
\hspace{1bp}Seq.~3\hspace{1bp}  &  298  &  98.3\%  &  97.8\%  &  51  &  99.3\%  &  98.9\%  &  61 \\
\hspace{1bp}Seq.~4\hspace{1bp}  &  371  &  99.7\%  &  99.4\%  &  53  &  99.6\%  &  99.7\%  &  45 \\
\hspace{1bp}Seq.~5\hspace{1bp}  &  298  &  99.0\%  &  98.7\%  &  41  &  99.9\%  &  99.5\%  &  50 \\
\hline
\end{tabular}
\normalsize
\end{center}
\caption{Results for respiratory gating on ultrasound images. For each
         image sequence, we show the number of frames it contains, the
         correlation coefficient~(CC) for kernel ridge regression~(KRR)
         and our sparse interpolator, and the number of support
         vectors~(SV's).
         Parameter values:
         $\lambda = 0.1$, $\varepsilon = 0.001$.}
 \label{tab:result} 
\end{table}

Respiratory gating tracks a patient's breathing cycle, which has numerous
applications such as 4D imaging, radiation therapy, and image
mosaicing~\cite{rohlfing2001modeling}. Manifold learning has been used for
highly accurate respiratory gating of ultrasound
images~\cite{wachinger2010miccai}, where 4D data reconstruction was achieved
with retrospective gating, i.e., the gating was calculated after the data
acquisition was finished. We extend this work to attain real-time gating. A
small number of breathing cycles are acquired and used as input for manifold
learning to construct the respiratory signal, as is done for retrospective
gating. The new incoming stream of ultrasound images is then gated by
performing an out-of-sample extension.

We conduct experiments on five 2D ultrasound image sequences of the human
liver acquired during free breathing; example images are shown in
Fig.~\ref{fig:USimages}. Each sequence contains 640$\times$480-pixel images
and vary in length between 298 and 371 frames captured at 33~Hz. For a given
image sequence, we use each image in the sequence as an input data point for
learning a 1D manifold with Laplacian eigenmaps~\cite{laplacian_lle}; we use
a 9-nearest-neighbor graph with an associated heat kernel of temperature
$t=10$. The 1D embedding learned using an entire sequence of images serves as
a reference signal for evaluating our sparse out-of-sample extension versus
kernel ridge regression as the baseline. In what follows, we compare the 1D
embedding of our sparse out-of-sample extension to the reference signal by
computing a correlation coefficient between them. We use kernel ridge
regression as a baseline method. Here we train on the first 200 frames and
test on the remaining frames. We then compare the results with those obtained
by training on all frames, as would be done for retrospective gating.

We first examine the influence of parameters $\varepsilon$ and $\lambda$ on
the resulting interpolator. Training on the first 200 images of one of the
ultrasound image sequences, we compute the correlation coefficient with the
reference signal and the number of support vectors versus the error tolerance
$\varepsilon$ (Fig.~\ref{fig:USimages}). As expected, smaller error tolerance
$\varepsilon$ requires more support vectors but also leads to a higher
correlation coefficient with respect to the reference signal. Also, a higher
kernel ridge regression regularization parameter~$\lambda$ leads to fewer
support vectors. However, stronger regularization also leads to lower
correlation coefficients. These results suggest a natural tradeoff between the
accuracy and the computational cost of the projection operation.

In the next experiment, we use $\lambda=0.1$ and $\varepsilon=0.001$. Training
on the first 200 frames and testing on the rest of the frames, we report the
correlation coefficients and the number of support vectors in
Table~\ref{tab:result}. The number of support vectors for kernel ridge
regression is 200 in this case. We then repeat the experiment, training on all
the frames. In this case, the number of support vectors for kernel ridge
regression is the length of the sequence. We achieve a high correlation for
all sequences, with a comparable performance between our sparse interpolator
and kernel ridge regression. Comparing the number of support vectors when
training on the first 200 frames vs.~training on all the frames, we note that
the number of support vectors stays roughly the same for a given image
sequence. This again suggests that the number of support vectors depends on
the low-dimensional embedding's complexity and not the training set size.

\subsection{Patient Position Estimation Using MRI}
\label{sec:mri}

\begin{figure}[t]
\begin{center}
\includegraphics[width=0.95\linewidth]{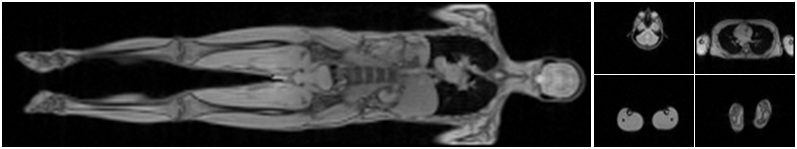}
\caption{Left: Coronal plane of MRI scan showing the entire patient.
         Right: Axial slices on which manifold learning is performed. 
\label{fig:mriEnt}}
\end{center}
\end{figure}

\begin{figure}[t]
\begin{center}
\subfloat[][\label{fig:MRa}] {
\includegraphics[width=0.315\linewidth, clip=true, trim=1.8in 3.5in 1.7in 3.5in]{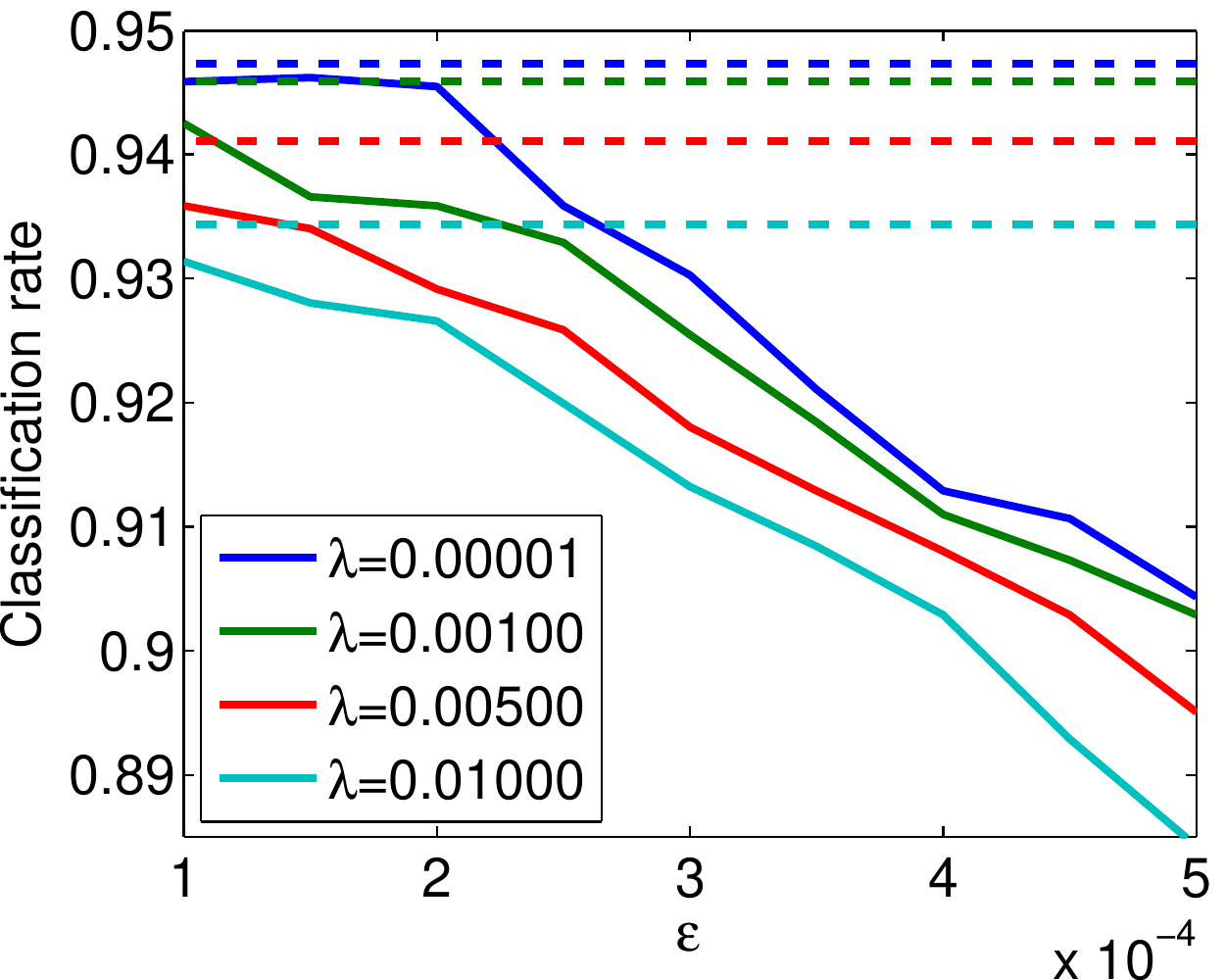}
}
\subfloat[][\label{fig:MRb}] {
\includegraphics[width=0.315\linewidth, clip=true, trim=1.8in 3.5in 1.7in 3.5in]{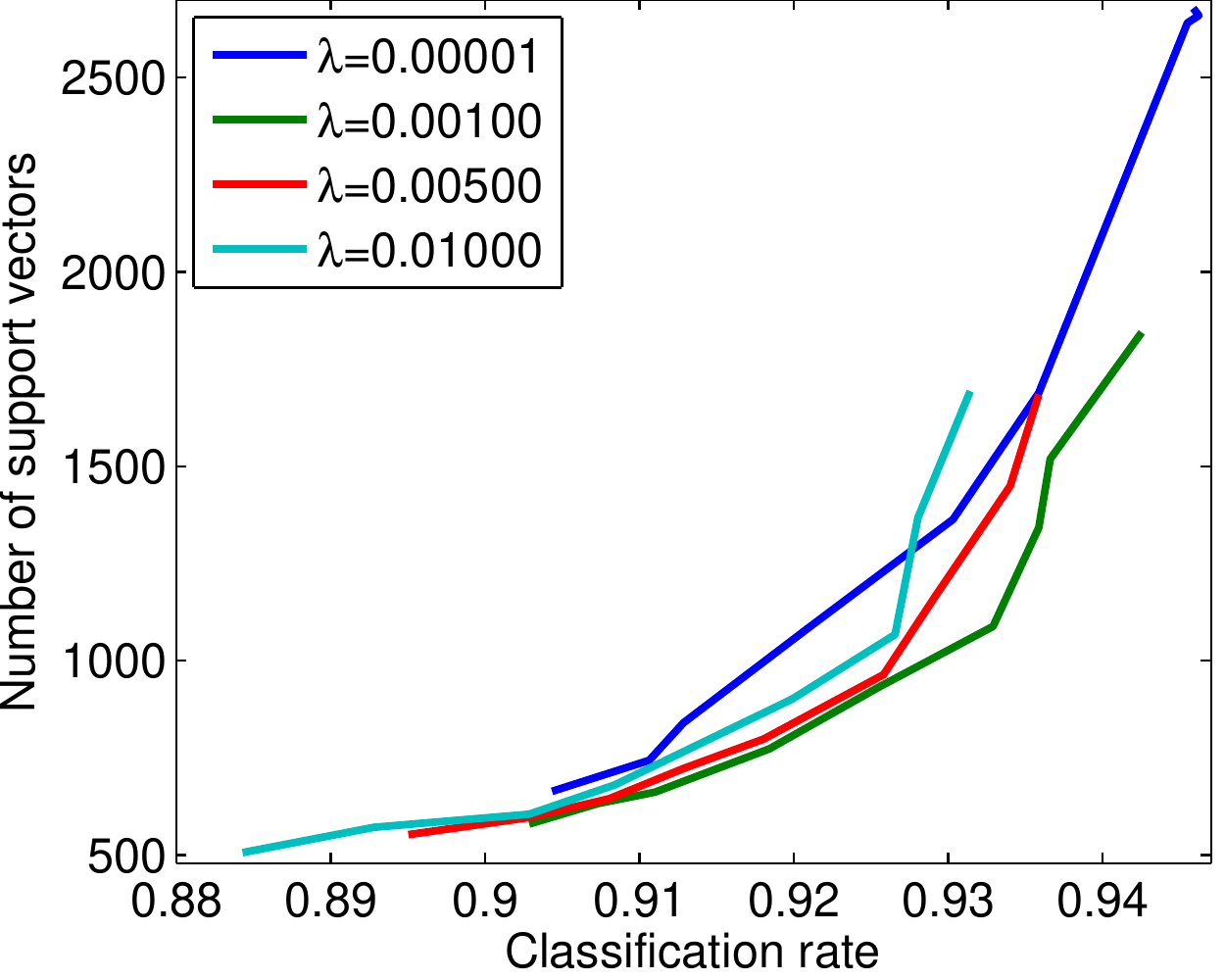}
}
\subfloat[][\label{fig:MRc}] {
\includegraphics[width=0.315\linewidth, clip=true, trim=1.8in 3.5in 1.7in 3.5in]{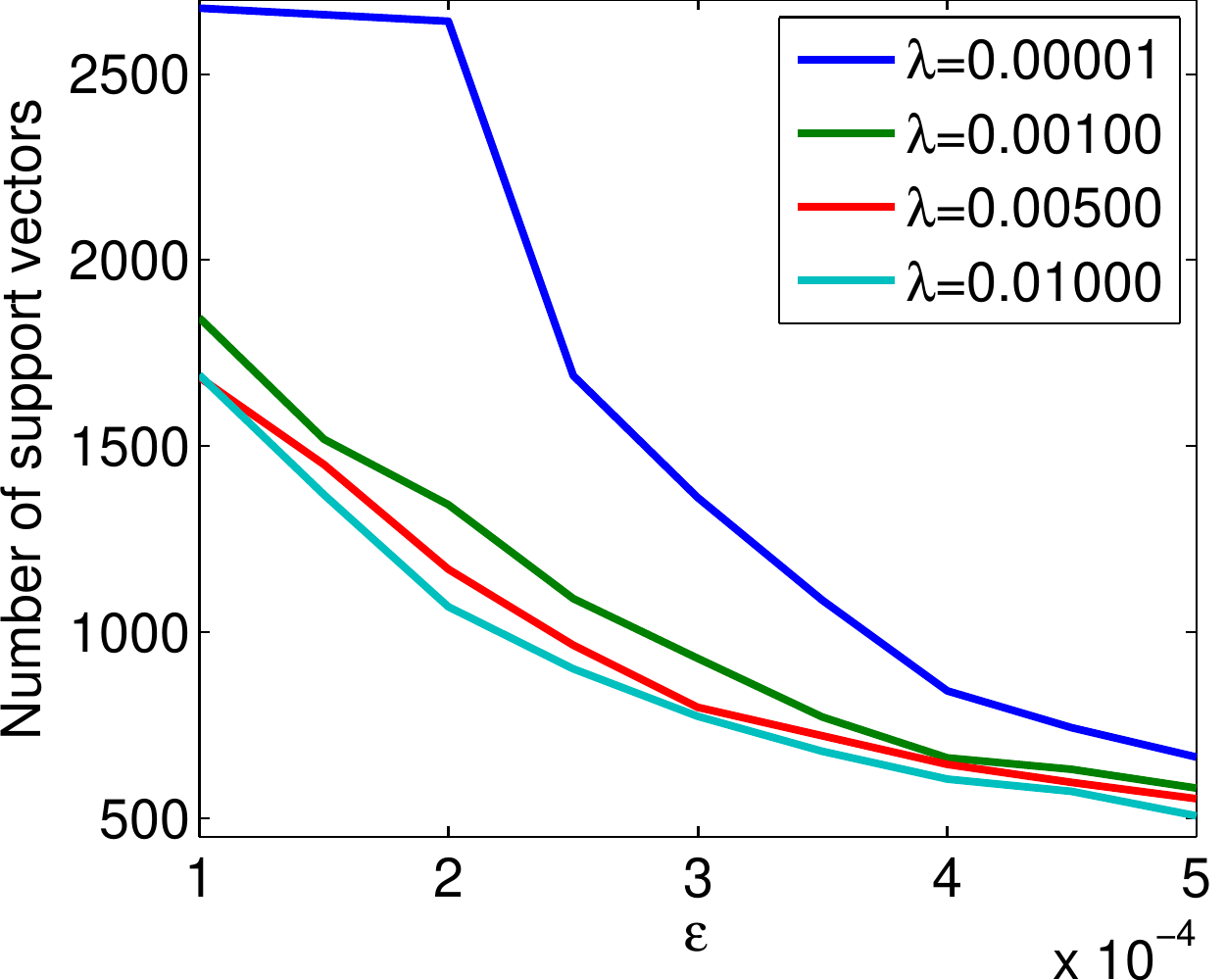}
}
\caption{Leave-one-out classification results for MRI data:
         (a) classification rate vs.~$\varepsilon$ for
             our sparse out-of-sample extension (solid line)
             and kernel ridge regression (dotted line)
         (b) the number of support vectors vs.~classification rate;
         (c) the number of support vectors vs.~error tolerance $\varepsilon$.
         All figures report results for different values of kernel ridge
         regression regularization parameter $\lambda$.
\label{fig:mriClass}}
\end{center}
\end{figure}

The radio frequency power in magnetic resonance imaging leads to tissue
heating and has to be monitored by measuring the specific absorption rate,
which depends on the position of the patient in the scanner. For current
high-resolution scanners, this imposes restrictions because either fewer
slices can be acquired or the in-plane resolution has to be reduced. Manifold
learning can be used to estimate the position of the patient in the
scanner~\cite{wachinger2010mani}.

First, low-resolution images are acquired while the bed that the patient lies
on moves inside the scanner. The images are embedded in a low-dimensional
space, where each axial image is associated with a body part (head, neck,
lung, etc.) using a nearest-neighbor classifier. By knowing which slices
correspond to which body parts, we can estimate the position of the patient in
the scanner. It is important that the estimation be done in real-time to
provide the position information before the high-resolution scan starts. In
this application, we can apply manifold learning offline on a large database
of scans. Then during the actual scan, we use an out-of-sample extension to
project the acquired slices into the low-dimensional space. For large training
datasets, it may be difficult to meet the time requirements with kernel ridge
regression. Consequently, the reduction to a small set of support vectors
offers a substantial advantage.

We run experiments on 13 whole body scans, such as the example shown in
Fig.~\ref{fig:mriEnt}. A medical expert assigned an anatomical label (head,
neck, lung, abdomen, upper leg, and lower leg) to each of the axial slices
(64$\times$64 pixels). We apply Laplacian eigenmaps to embed the high
dimensional slices in a two-dimensional space; we use a 40-nearest-neighbor
graph with a heat kernel of temperature $t=49$. To predict the anatomical
label of an axial image, we perform nearest-neighbor classification in the
learned low-dimensional space. We repeat this classification procedure for
different values of error tolerance $\varepsilon$ ranging from
$1\times10^{-4}$ to $5\times10^{-4}$.

We compare the classification performance of embeddings obtained from
our sparse interpolator and kernel ridge regression.
Fig.~\ref{fig:mriClass}\subref{fig:MRa} reports leave-one-out classification
performance for different values of error tolerance $\varepsilon$. The
classification rates for kernel ridge regression are provided for comparison;
they do not change for different values of~$\varepsilon$.
Figs.~\ref{fig:mriClass}\subref{fig:MRb}
and \ref{fig:mriClass}\subref{fig:MRc} characterize the sparsity of the
interpolation function constructed by reporting the number of support vectors
as a function of the classification rate and error tolerance~$\varepsilon$.
The total number of frames used in this experiment is 2697, which corresponds
to the number of support vectors for kernel ridge regression. We observe a
clear correlation between error tolerance~$\varepsilon$ and the classification
performance. Smaller values of $\varepsilon$ lead to better classification
performance but require more support vectors. Thus, we can trade off
computational speed with classification performance by tuning parameters
$\lambda$ and $\varepsilon$ to be as large as possible while maintaining a
classification rate above a minimum tolerated threshold.

\section{Conclusion}
\label{sec:conclusion}

We derived a novel method for multivariate regression that approximates kernel
ridge regression, where the final estimated interpolation function depends
only on a subset of the original input points acting as support vectors. Our
approach provides a guarantee on the approximation error for training data.
We applied our method as an out-of-sample extension for manifold learning,
illustrating applications to respiratory gating and MRI classification. 

Turning toward nonlinear dimensionality reduction more generally, many widely
used algorithms are computationally expensive for massive datasets. Thus,
ideally we would like to find support vectors first, before applying
dimensionality reduction. Our results suggest that the support vectors for
interpolation may not be uniformly sampled in the input space. This invites
the question of how to non-uniformly sample training data in the input space
and adjust a dimensionality reduction algorithm accordingly to account for the
geometry of these~samples.

\textbf{Acknowledgements.}
We thank Siemens Healthcare for image data. This work was funded in part by
the National Alliance for Medical Image Computing (grant NIH NIBIB NAMIC
U54-EB005149) and the National Institutes of Health (grants NIH NCRR NAC
P41-RR13218 and NIH NIBIB NAC P41-EB-015902).

\renewcommand\refname{References}
\bibliographystyle{splncs03}
\bibliography{l1l2_sparsify}

\begin{thebibliography}{10}
\providecommand{\url}[1]{\texttt{#1}}
\providecommand{\urlprefix}{URL }

\bibitem{rk_theory}
Aronszajn, N.: Theory of reproducing kernels. Trans. AMS  (1950)

\bibitem{sparse_optimization}
Bach, F.R., Jenatton, R., Mairal, J., Obozinski, G.: Optimization with
  sparsity-inducing penalties. Foundations and Trends in Machine Learning
  (2012)

\bibitem{fista}
Beck, A., Teboulle, M.: A fast iterative shrinkage-thresholding algorithm for
  linear inverse problems. SIAM Journal on Imaging Sciences  (2009)

\bibitem{laplacian_lle}
Belkin, M., Niyogi, P.: Laplacian eigenmaps and spectral techniques for
  embedding and clustering. In: NIPS (2002)

\bibitem{out_of_sample_ext}
Bengio, Y., Paiement, J.F., Vincent, P., Delalleau, O., Roux, N.L., Ouimet, M.:
  Out-of-sample extensions for {LLE}, {I}somap, {MDS}, eigenmaps, and spectral
  clustering. In: NIPS (2004)

\bibitem{BhatiaRPWHR12}
Bhatia, K.K., Rao, A., Price, A.N., Wolz, R., Hajnal, J.V., Rueckert, D.:
  Hierarchical manifold learning. In: MICCAI (2012)

\bibitem{rrkrr}
Cawley, G.C., Talbot, N.L.C.: Reduced rank kernel ridge regression. Neural
  Processing Letters  (2002)

\bibitem{hessian_lle}
Donoho, D.L., Grimes, C.: Hessian eigenmaps: New locally linear embedding
  techniques for high-dimensional data. PNAS  (2003)

\bibitem{svr}
Drucker, H., Burges, C.J.C., Kaufman, L., Smola, A.J., Vapnik, V.: Support
  vector regression machines. In: NIPS (1997)

\bibitem{georg08manifold4D}
Georg, M., Souvenir, R., Hope, A., Pless, R.: Manifold learning for 4d ct
  reconstruction of the lung. In: CVPR Workshops (2008)

\bibitem{gerber09}
Gerber, S., Tasdizen, T., Joshi, S., Whitaker, R.: On the manifold structure of
  the space of brain images. In: MICCAI (2009)

\bibitem{hamm09}
Hamm, J., Davatzikos, C., Verma, R.: Efficient large deformation registration
  via geodesics on a learned manifold of images. In: MICCAI (2009)

\bibitem{dimensionality_reduction_review}
van~der Maaten, L.J.P., Postma, E.O., van~den Herik, H.J.: Dimensionality
  reduction: A comparative review. Tilburg University Technical Report  (2008)

\bibitem{l0_nphard}
Natarajan, B.K.: {Sparse Approximate Solutions to Linear Systems}. SIAM J.
  Comput.  (1995)

\bibitem{rohdeWPM08}
Rohde, G.K., Wang, W., Peng, T., Murphy, R.F.: Deformation-based nonlinear
  dimension reduction: Applications to nuclear morphometry. In: ISBI (2008)

\bibitem{rohlfing2001modeling}
Rohlfing, T., Maurer, Jr., C.R., O'Dell, W.G., Zhong, J.: Modeling liver motion
  and deformation during the respiratory cycle using intensity-based free-form
  registration of gated {MR} images. In: Medical Imaging: Visualization,
  Display, and Image-Guided Procedures (2001)

\bibitem{lle}
Roweis, S.T., Saul, L.K.: Nonlinear dimensionality reduction by locally linear
  embedding. Science  (2000)

\bibitem{krr}
Saunders, C., Gammerman, A., Vovk, V.: Ridge regression learning algorithm in
  dual variables. In: ICML (1998)

\bibitem{learning_with_kernels}
Sch\"{o}lkopf, B., Smola, A.J.: Learning with Kernels: Support Vector Machines,
  Regularization, Optimization, and Beyond. MIT Press (2001)

\bibitem{suzuki2010massive}
Suzuki, K., Zhang, J., Xu, J.: Massive-training artificial neural network
  coupled with laplacian-eigenfunction-based dimensionality reduction for
  computer-aided detection of polyps in ct colonography. IEEE TMI  (2010)

\bibitem{isomap}
Tenenbaum, J.B., de~Silva, V., Langford, J.C.: A global geometric framework for
  nonlinear dimensionality reduction. Science  (2000)

\bibitem{wachinger2010mani}
Wachinger, C., Mateus, D., Keil, A., Navab, N.: Manifold learning for patient
  position detection in {MRI}. In: ISBI (2010)

\bibitem{wachinger2010miccai}
Wachinger, C., Yigitsoy, M., Navab, N.: Manifold learning for image-based
  breathing gating with application to {4D} ultrasound. In: MICCAI (2010)

\bibitem{zhang06}
Zhang, Q., Souvenir, R., Pless, R.: {On Manifold Structure of Cardiac MRI Data:
  Application to Segmentation}. CVPR  (2006)

\end{thebibliography}

\end{document}